\documentclass[sigconf]{acmart}
\AtBeginDocument{%
  }

\setcopyright{acmlicensed}
\copyrightyear{2025}
\acmYear{2025}
\acmDOI{XXXXXXX.XXXXXXX}
\acmConference[Conference acronym 'XX]{Make sure to enter the correct
  conference title from your rights confirmation email}{June 03--05,
  2018}{Woodstock, NY}
\acmISBN{978-1-4503-XXXX-X/2018/06}

\usepackage{amsmath}
\usepackage{cancel}
\usepackage{algorithm,algpseudocode}
\usepackage{refcount}

\begin{document} 

\title{Heterogeneous Causal Learning for Optimizing Aggregated Functions in User Growth} 

%

%

\author{Shuyang Du}
\email{shuyangdu139@gmail.com} 
\affiliation{%
 \institution{Lightspeed}
 \streetaddress{2200 Sand Hill Road}
 \city{Menlo Park}
 \state{CA}
 \country{USA}
}

\author{Jennifer Y. Zhang} 
\email{jenniferyt.zhang@mail.utoronto.ca} 
\affiliation{%
 \institution{University of Toronto}
 \streetaddress{27 King\'s College Cir, Toronto}
 \city{Toronto}
 \state{ON}
 \country{CA}
}

\author{Will Y. Zou}
\email{will@angle.ac}
\affiliation{%
 \institution{Angle.ac}
 \city{San Francisco}
 \state{CA}
 \country{USA}
}

%

\begin{abstract} 
User growth is a major strategy for consumer internet companies. To optimize costly marketing campaigns and maximize user engagement, we propose a novel treatment effect optimization methodology to enhance user growth marketing. By leveraging deep learning, our algorithm learns from past experiments to optimize user selection and reward allocation, maximizing campaign impact while minimizing costs. Unlike traditional prediction methods, our model directly models uplifts in key business metrics. Further, our deep learning model can jointly optimize parameters for an aggregated loss function using softmax gating. Our approach surpasses traditional methods by directly targeting desired business metrics and demonstrates superior algorithmic flexibility in handling complex business constraints. Comprehensive evaluations, including comparisons with state-of-the-art techniques such as R-learner and Causal Forest, validate the effectiveness of our model. We experimentally demonstrate that our proposed constrained and direct optimization algorithms significantly outperform state-of-the-art methods by over $20\%$, proving their cost-efficiency and real-world impact. The versatile methods can be applied to various product scenarios, including optimal treatment allocation. Its effectiveness has also been validated through successful worldwide production deployments.


\end{abstract} 






\maketitle 

\section{Introduction}
\label{sec:intro} 
Enhancing user growth and engagement is a critical focus for internet companies operating in rapidly evolving markets. As the cost of user acquisition continues to rise, products are increasingly turning to targeted marketing campaigns and cross-selling strategies to drive growth. Industries such as ride-sharing (Uber, Lyft), accommodation (Airbnb), and e-Commerce (Amazon, eBay) employ various strategies to incentivize user behavior and foster loyalty.

As demonstrated in previous research \cite{halperin2018toward, cohen2018frustration}, offering rewards to users, even without explicit apologies for negative experiences, can have a positive impact on future engagement and revenue. These findings underscore the potential of targeted marketing interventions to influence user behavior. It is important to note that while such strategies drive growth, they can incur significant costs. To optimize the allocation of resources, it is essential to adopt a rigorous data-driven approach that directly targets the causal impact of treatments on business outcomes.

Traditional methods, such as churn prediction and rule-based methods, often focus less on capturing user behavior or the long-term effects of marketing interventions. A more modern approach is the treatment effect estimation~\cite{rubin1974estimating} framework for understanding the causal impact of treatments on user behavior. We enhance the treatment effect functional forms in this framework to propose deep learning algorithms which identify optimal strategies for allocating rewards and incentives to maximize overall business objectives while minimizing costs.

Our work aims to develop a business decision methodology that optimizes treatment effectiveness, minimizing costs while maximizing user engagement. Key contributions of this paper include:  

\begin{itemize}

\item \textbf{Heterogeneous Treatment Effect-based Business Decisions} - Our method makes decisions on the heterogeneous treatment effects of key business metrics, making it adaptable to various business use cases with minimal customization. This contrasts with traditional approaches which rely on heuristic rules or models tailored to specific scenarios. This approach also enables effective evaluation and provides actionable insights for data-driven decision making.

\item \textbf{Benefit vs. Cost for Aggregated Efficiency} - 
Our approach optimizes the efficiency ratio of value to cost ($\delta_{\text{value}}/\delta_{\text{cost}}$) to address the need of real-world applications, while most existing research focuses on the treatment effect of a single outcome. We go beyond point estimates to optimize for the \emph{aggregated treatment effect functions}, providing a more comprehensive and robust decision-making framework to solve multiple challenges together.

\item \textbf{Deep Learning Integration and Joint Objective} - Our approach aggregates multiple outcomes into a unified learning objective, allowing the optimization across all parameters through deep learning. Previous methods focus on estimating treatment effects for individual outcomes in a greedy manner, relying on statistical or linear models. Integration with deep learning enhances the flexibility and scalability of our methodology. 

\item \textbf{Barrier Function for Constrained Optimization} - Real-world constraints, such as budget limitations and geographic restrictions, significantly influence model results. We propose a constrained ranking algorithm that considers the impact of actions and constraints in production environments. This versatile model addresses real-world needs and limitations, while optimizing for market-wide efficiency.
\end{itemize} 

\section{Background} 
\label{sec:related_work} 
Previous research has explored methods to optimize user marketing, rewards, and retention. \citet{halperin2018toward} and \citet{cohen2018frustration} studied the impact of apology treatments on the recovery of user trust. \citet{andrews2016mobile} examined factors influencing coupon redemption, while \citet{hanna2016optimizing} and \citet{Manzoor2017RUSHTT} investigated factors affecting the redemption of time-limited incentives. Churn prediction methods~\cite{vafeiadis20151,lalwani2022customer,ullah2019churn} used boosting, SVM, and classification techniques to directly predict user behavior. These studies focused primarily on direct prediction or classification and the average treatment effect of redemption or exploration instead of user selection and optimization. 

While the aforementioned methods addressed the business problem, they did not fully leverage causal learning principles. Causal inference framework by ~\citet{rubin1974estimating} provided a foundation for studying treatment effects. User instances were treated with specific actions, and the observed outcomes were used to fit models. A significant approach involved statistical methods such as \emph{meta-learners}~\cite{kunzel2017meta},  decomposing the learning algorithms into composite models. Another approach used decision trees and random forests~\cite{chen2016xgboost}, such as uplift trees~\cite{rzepakowski2012decision}, causal  forests~\cite{wager2017estimation,athey2016recursive}, boosting~\cite{powers2017some}, and latent variable models~\cite{louizos2017causal} to build powerful causal inference models.

Recently, quasi-oracle estimation~\cite{nie2017quasi} has emerged as a popular framework for learning heterogeneous treatment effects. It effectively estimates treatment effects for single outcomes by considering both \emph{Conditional Treatment Effect (CTE)} and \emph{Average Treatment Effect (ATE)}. However, these algorithms are limited to single-outcome scenarios and cannot handle multiple outcomes or benefit-cost trade-offs.

In this work, we propose a set of algorithms that not only predict treatment effects but also combine multiple outcomes into comprehensive effectiveness measures that can be jointly optimized.

\paragraph{\textbf{Notations for Treatment Effect Estimation}} As background to present the proposed algorithms, we review the potential outcomes framework \cite{neyman23thesis,rubin1974estimating,nie2017quasi} to estimate treatment effects. The framework considers $n$ independent and identically distributed subjects. For a single subject $i$, let $\mathbf{x}^{(i)}$ represent its feature vector. Let $\mathbf{X}$ denote the collection of feature vectors in the entire set of subjects. 

In experiments, treatments are assigned to subjects, and outcomes are observed. The outcomes are denoted as $Y_1^{(i)}$ and $Y_0^{(i)}$,  representing the observed outcome under treatment and control conditions, respectively. The binary treatment assignment $T^{(i)}$ indicates whether user $i$ received the treatment ($T^{(i)}=1$) or not ($T^{(i)}=0$). The treatment propensity, $e(\mathbf{x}^{(i)})$, represents the probability of a user receiving treatment given their features $\mathbf{x}^{(i)}$, i.e. $e(\mathbf{x}^{(i)}) = P (T = 1  | \mathbf{x}^{(i)})$. 

The framework aims to estimate the treatment effect function conditioned on user features $\mathbf{x}$, i.e. $\tau^*(\mathbf{x}) = E(Y_1 - Y_0 | \mathbf{x})$. The historical experimental data serves as the foundation for model fitting. Once fitted, the model can be used to predict treatment decisions on new data. 

For all causal models in this paper, we follow the unconfounded assumption. The potential outcomes are independent of the treatment assignment given the observed covariates \cite{Rosenbaum83propensity}: $\{Y_0, Y_1\} \perp T^{(i)}|\mathbf{x}^{(i)}$. 

In a problem with treatment effect uplifts on revenue and cost, we consider the notation:
\begin{align*}
    \tau^{*r}(\mathbf{X}) = E(Y^r_1 - Y^r_0 | \mathbf{X}) \quad \text{and} \quad
    \tau^{*c}(\mathbf{X}) = E(Y^c_1 - Y^c_0 | \mathbf{X}).
\end{align*}

These are the conditional treatment effect functions across the set of users, which are the functional forms we will leverage in deep learning algorithms. 

\begin{figure*}
  \centering 
  \includegraphics[width=1.0\textwidth]{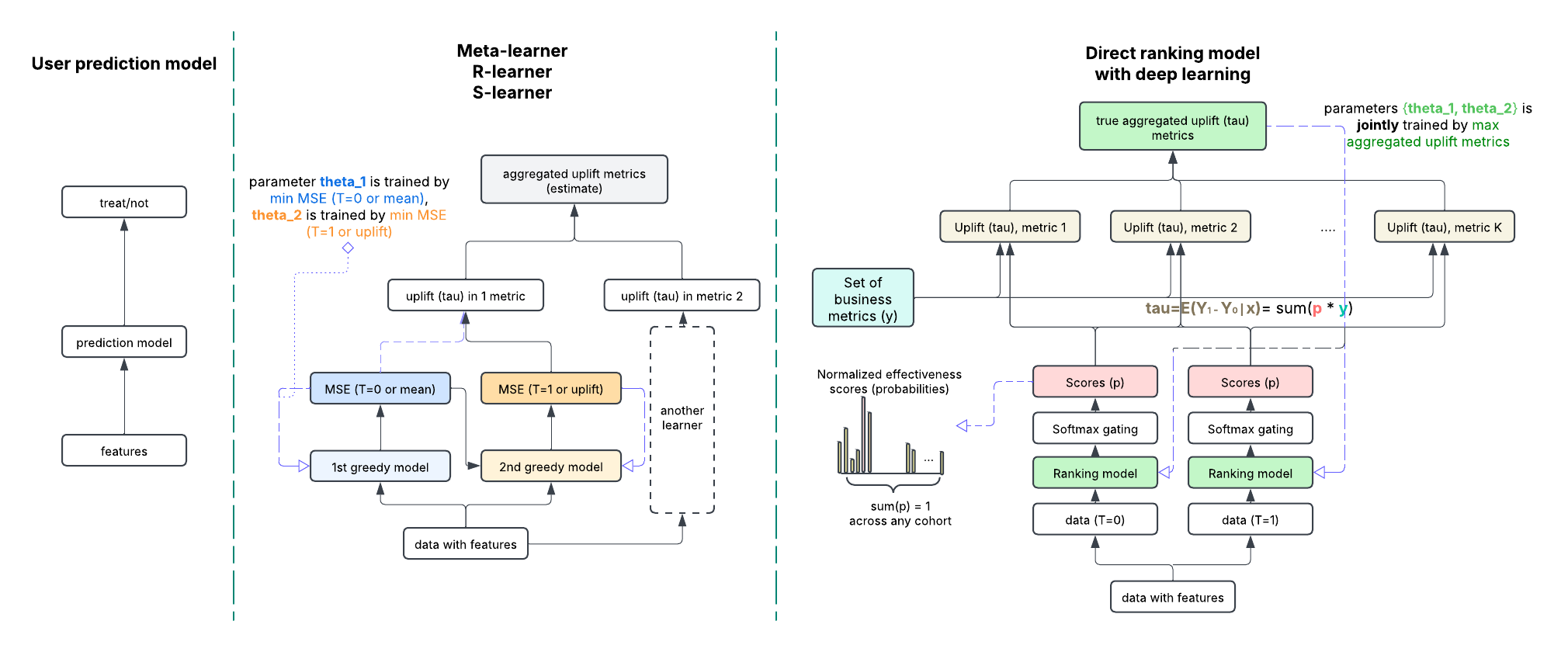} 
  \caption{Graph comparison of prior work and DRM models.} 
  \label{fig:algorithm_illustration} 
\end{figure*} 

\textbf{Quasi-Oracle Estimation (R-learner)}: The quasi-oracle estimation algorithm~\cite{nie2017quasi} estimates treatment effects using a two-step process. It first estimates the expected outcome without treatment and the probability of receiving treatment (propensity score). This helps separate out the part of the data related to the treatment effect. It then uses regression to estimate the treatment effect (uplift).

Concretely, the conditional mean of outcomes given user features are $\mu^{*}_T(\mathbf{x}) = E(Y_T | \mathbf{x})$. The expected outcome under treatment indicator $T^{(i)}$ is $E(Y_T^{(i)} | \mathbf{x}^{(i)}) =  \mu^*_0(\mathbf{x}^{(i)}) + T^{(i)}\tau^*(\mathbf{x}^{(i)})$, where \(\mu^*_0(\mathbf{x}^{(i)})\) is the expected outcome without treatment, and \(\tau^*(\mathbf{x}^{(i)})\) is the treatment effect we aim to estimate.

We define the error term as
\begin{align*}
\epsilon(T^{(i)}) = Y_T^{(i)} - E(Y_T^{(i)}) = Y_T^{(i)} - (\mu^*_0(\mathbf{x}^{(i)}) + T^{(i)}\tau^*(\mathbf{x}^{(i)})).
\end{align*}
We also have
\begin{align*}
m^*(\mathbf{x}^{(i)}) = E(Y^{(i)} | \mathbf{x}^{(i)}) = \mu^*_0(\mathbf{x}^{(i)}) + e^*(\mathbf{x}^{(i)})\tau^*(\mathbf{x}^{(i)}),
\end{align*}
where $e^*(\mathbf{x}^{(i)})$ is the probably of use $i$ receiving treatnent (propensity score).

Substituting $\mu^*_0(\mathbf{x}^{(i)})$, we arrive at the decomposition\footnote{At the end of the equation, the expected value of the error $\epsilon$ across data and expected value of Y is $0$ given unconfoundedness assumption: $E(\epsilon(T^{(i)}) | \mathbf{X}^{(i)}, T^{(i)}) = 0$.}: 
\begin{equation}
\label{eq:quasi_oracle_balance}
Y^{(i)} - m^*(\mathbf{x}^{(i)}) = (T^{(i)} - e^*(\mathbf{x}^{(i)}))\tau^*(\mathbf{x}^{(i)}) + \epsilon.
\end{equation}

In practice, R-learner first uses regression to fit the $m^*$ (the expected outcome given features) and $e^*$ (propensity) models. The prediction results of the regression are then used to estimate the treatment effect, $\tau^*$ model. After learning, $\tau^*$ function can be used to estimate the treatment effect given user with feature $\mathbf{x}^{(i)}$. 

In a problem with multiple uplifts or treatment effects (e.g. $\tau^r$ for revenue, and $\tau^c$ for cost), the algorithm is fitted greedily. The resulting model can be used to separately estimate $\tau^r$ and $\tau^c$. 

In addition, we refer to the Meta-learners study~\cite{kunzel2017meta} and churn prediction methods~\cite{vafeiadis20151,lalwani2022customer,ullah2019churn}. Along with R-learner, the Meta-learners are effective for estimating conditional treatment effects. However, both models are composed of distinct parts with its own set of parameters. These part models are learned individually and greedily on the training data. Moreover, R-learner and Meta-learner models focus on the treatment effect of a single outcome or a single metric uplift, instead of an objected aggregated one with multiple uplift functions. Besides, while the prior work mainly uses a linear model, this paper extends the R-learner framework using a deep learning approach to enhance modeling capacity.

The separate-model-training and single-business-metric characteristics of R-learner and Meta-learners are illustrated in Fig.~\ref{fig:algorithm_illustration}. 

\section{Algorithms} 
\label{sec:algorithms} 

We propose novel approaches for treatment effect optimization by solving an aggregated objective, maximizing reward while minimizing cost. We define the uplift $\tau^{*r}$ and $\tau^{*c}$ as functional forms of the input. The parameters of these functions can be optimized  during training \textbf{jointly and non-greedily} with respect to an overall, aggregated objective that is constructed with $\tau^{*r}$ and $\tau^{*c}$. 


\subsection{Direct Ranking Model} 
\label{sec:drm}

We motivate the business objective to identify a \emph{portfolio of users} that we can achieve highest incremental user uplift on desired metrics, which does not rely on the perfect individual prediction (point estimate) of treatment effect, but achieves the overall market-wide effectiveness. We illustrate the model structure in Fig.~\ref{fig:algorithm_illustration}.


\textbf{Wholistic Optimization.} Without performing point estimates, we define the unconstrained optimization problem where we maximize the definition of Return on Investment (ROI) for user growth: 
\begin{align} 
\label{eq:pstatement_unconstrained_opt} 
\text{maximize} \quad \frac{\text{Incremental Value}}{\text{Incremental Cost}} = \frac{\tau^{*r}(\mathbf{\theta}, \mathbf{X})}{\tau^{*c}(\mathbf{\theta}, \mathbf{X})}.
\end{align} 

The objective function is an \emph{aggregation of treatment effect functions}. In prior work, this maximization of the aggregated treatment effected functions is not possible due to the lack of \emph{treatment effect functional forms of model parameters and the input}, namely, $\tau (\mathbf{\theta},\mathbf{x})$. 

\textbf{Functional Form for Treatment Effect.} The treatment effect function is an expected value of the outcome $Y$, and the $\tau$ function can be expressed with effectiveness probabilities and the observed outcome Y: 
\begin{align}
\label{eq:treatment_effect_functional_form}
\tau (\mathbf{\theta},\mathbf{X}) &= \mathbf{E}(Y_1-Y_0|\mathbf{X}, \mathbf{\theta}) \notag\\
&= \sum_{T_i=1} p_i(\cdot|\mathbf{\theta}, \mathbf{x}) Y^{(i)} - \sum_{T_i=0} p_i(\cdot|\mathbf{\theta}, \mathbf{x}) Y^{(i)}.
\end{align} 
The above form of $\tau (\mathbf{\theta},\mathbf{x})$ is expressed in terms of user effectiveness probabilities $p_i(\cdot|\mathbf{\theta}, \mathbf{x})$. If we are able to express $p_i(\cdot|\mathbf{\theta}, \mathbf{x})$ in functional form, then $\tau(\mathbf{\theta}, \mathbf{x})$ can be aggregated in Eq.~\ref{eq:pstatement_unconstrained_opt} to form the ROI maximization objective. 

\textbf{Softmax Network for User Effectiveness Probabilities.} To achieve this, we leverage the critical idea of the softmax network with deep learning. When applied across a user cohort, the softmax network normalizes model scores to sum to one, making them user effectiveness probabilities. The inspiration comes from dynamic routing in Mixture of Experts~\cite{shazeer2017outrageously}, language models~\cite{brown2020language}, and search ranking~\cite{huang2013learning,valizadegan2009learning}, where softmax is used to normalize scores for selecting instances across a population. In our algorithm, due to its versatility, the softmax can be applied to control, treatment, and other cohorts of users. Importantly, softmax retains the \emph{ranking order} of instances after normalization, so it is ideal for user ranking. We note that the \emph{user effectiveness probabilities} is not a measure of how likely the selection of a user will maximize the uplift of one business metric, but a measure for how likely this selection will maximize the overall objective composed of \emph{aggregated treatment effect functions} defined in Eq.~\ref{eq:pstatement_unconstrained_opt} and we propose the corresponding DRM model in Fig.~\ref{fig:algorithm_illustration}. Formally, the user effectiveness probabilities are defined as: 
\begin{align}
\label{eq:softmax_effectiveness_probabilities}
p(.|\mathbf{\theta}, \mathbf{x}) = \text{softmax}(s)
=\frac{\exp(f(\mathbf{x}|\mathbf{\theta}))}{\sum_{\text{cohort}}\exp(f(\mathbf{x}|\mathbf{\theta}))},
\end{align}
where the scores $s$ are predicted by a model, and the functional form $f$ can be expressed as a neural network\footnote{Commonly, we use $\tanh$ as the last non-linearity operator of this neural network.}: $s=f(\mathbf{x} | \mathbf{\theta})$. 

\textbf{Optimizing With Deep Learning}. Putting together Eq.~\ref{eq:pstatement_unconstrained_opt}, Eq.~\ref{eq:treatment_effect_functional_form}, and Eq.~\ref{eq:softmax_effectiveness_probabilities} , the learning objective is given in Eq.~\ref{eq:drm_deep_learning_summary}, which includes a regularization term for $f$.
\begin{equation}
  \label{eq:drm_deep_learning_summary}
  \hat{\theta} = argmax_{\theta}\left \{ \frac{\tau^{*r}(\theta, \mathbf{X})}{\tau^{*c}(\theta, \mathbf{X})}-\Lambda_n(f(\cdot ))  \right \}.
\end{equation}
The parameters defining the function $f$ can be trained by maximizing the objective function in deep learning. For broader use cases, various forms of $f$ function can be used, such as multi-layer perceptrons, convolution, or recurrent neural networks. 

The $\tau$ terms are: 
\begin{equation}
  \label{eq:tau_detailed_forms}
  \tau^{*r}=\sum_{i=1}^nY^{r(i)}p_i(\mathbb{I}_{T_i=1} - \mathbb{I}_{T_i=0}), \quad \tau^{*c}=\sum_{i=1}^nY^{c(i)}p_i(\mathbb{I}_{T_i=1} - \mathbb{I}_{T_i=0}),
\end{equation}
where $\mathbb{I}$ is the indicator function, equal to $1$ if the sub-scripted condition is met, and $0$ otherwise.

As with prior work for estimating treatment effect, we adopt the explore-exploit framework for model training. The training data is first collected by applying treatment to users, and outcomes are observed. The model can be trained with objective in Eq.~\ref{eq:drm_deep_learning_summary}. After training, the scoring function $f$ is used to score, select, and rank users on new data. 

\textbf{Working with Propensity Functions}. For experiments where treatment isn't given randomly, propensity~\cite{rubin1974estimating,nie2017quasi} ensures proper weighting to address biases. We offer a full derivation of the functional form of the treatment effect function with a propensity function in Appendix~\ref{apB}. 

In our algorithm, the propensity function $e(\mathbf{x}) = P(T=1 | \mathbf{x})$ has its standard form similar to prior art, and it can be estimated given the training data. We offer the form of the treatment effect function and its relationship with the propensity function $e(x)$:
\begin{equation}
\label{eq:treament_effect_function_with_propensity}
\tau^{*} = \hat{e} \sum\limits_{i=1}^{n} \frac{1}{e(\mathbf{x}^{(i)})} Y^{(i)} p_i \mathbb{I}_{T_i=1} - (1-\hat{e}) \sum\limits_{i=1}^{n} \frac{1}{1-e(\mathbf{x}^{(i)})} Y^{(i)} p_i \mathbb{I}_{T_i=0},
\end{equation}
where $\hat{e}$ is the number of treated instances over the total number of instances. 

DRM propensity weighted objective can be built by combining Eq.~\ref{eq:treament_effect_function_with_propensity} into Eq.~\ref{eq:drm_deep_learning_summary}. In experiments, we find Direct Ranking Model with propensity weighting is useful for interpreting counterfactuals, discussed in Section~\ref{sec:empirical_results}\footnote{To apply propensity in DRM, we note that the training and prediction data assumptions as well as their measurement metrics should align.}.

\subsection{Constrained Ranking Model} 
Constraints such as a fixed budget are inherent in growth products and they can bring challenges to the optimization~\cite{marquez2017imposing}.

To address this, we transform hard constraints into soft constraints similar to barrier methods~\cite{norcedal2006numerical}. When the constraints are known beforehand, the model optimizes its parameters to best solve the better-defined constrained problem, improving model performance. 

\textbf{Barrier Function}. Building on the direct ranking approach in \ref{sec:drm}, we define barrier functions which modify the objective function to include constraints. The barrier function penalizes the model unless the desired fraction of users is selected. The objective with a barrier function can be computed given any batch of data, and the barrier function is re-computed for every mini-batch iteration during optimization. We consider either a percentage $P$ or a cost budget $B$ constraint.

We illustrate the solution procedure below. Let $\mathbf{p}=\{p_{1}, p_{2}, ..., p_{N}\}$ denote the collection of normalized user effectiveness probabilities, where each $p_{i}$ is given in Eq.~\ref{eq:softmax_effectiveness_probabilities}. Given a batch of users, we create a barrier function to zero out the effective probabilities that violate the constraints:
\begin{itemize}
    \item Sort the probabilities $\mathbf{p}=\{p_{1}, p_{2}, ..., p_{N}\}$ into $\{p_{k_1}, p_{k_2}, ..., p_{k_N}\}$, where each $k_j$ is a sorted index.
    \item Determine offset $d^*$ as a threshold\footnote{Or as a value in the in-between interval of two probabilities.} in the set $\mathbf{p}_s$ that defies the location of the barrier. 
    \item With $T$ as temperature hyperparameter to control the softness of the barrier, define the barrier function as: $\sigma(x) = \text{sigmoid}(-T (x - d^{*}))$.
    \item Multiply each user probability $p_{i}$ with its barrier function, i.e. $\hat{p}_{i} = p_{i}\sigma(p_{i})$. 
    This allows the algorithm to keep each $p_{i}$ if it is above the barrier \(d^{*}\) and nullify $p_{i}$ if it is below the barrier.
    \item Re-normalize $\hat{p}_{i}$ with softmax function so that they sum to one within the user cohort.
    \item Use the re-normalized probabilities to compute expectations in Eq.~\ref{eq:tau_detailed_forms}.
\end{itemize}

The algorithm can be adapted to various forms of constraints. In this paper, we offer two constraint definitions: 

\begin{itemize}
    \item \emph{Percentage Constraint}: sort users according to their effectiveness probabilities $\mathbf{p}$, and choose $d^{*}$ such that the top $P$ percent of users are selected. 
    \item \emph{Cost Budget Constraint}: sort users according to their corresponding cost values, aggregate the cumulative costs, and choose $d^*$ along the cumulative sum list, such that the cost budget is closely reached and not exceeded. 
\end{itemize}

In practice, we replace the normalized effectiveness probabilities $p$ in Eq. \ref{eq:softmax_effectiveness_probabilities} with the probability after pooling $\hat{p}$, then re-normalize it again afterwards. The model can be trained end-to-end with gradient methods. The barrier method can be seen as a dynamic form of pooling~\cite{jarrett2009best}. The model dynamically creates connection patterns in the neural network to focuses on the largest activations. In every optimization iteration, the objective with barrier function measures valid users according to constraints.

\textbf{Annealing}. The temperature term $T$ in the barrier function determines the sharpness of the selection method. We observed difficulties of optimizing the model with constrained ranking when $T$ is set large. The optimization algorithm could not find local minima. 

To address this issue, we propose an annealing process on the parameter $T$ to have a schedule of rising temperature. This allows the model to explore and settle into better local optima early in training while tightening constraints in later stages. The annealing process facilitates the optimizer to find an optimal solution. 


\subsection{Duality R-learner} 

In addition to the deep learning models, we develop a constrained optimization algorithm on top of R-learner. This Duality R-learner method is used to compare with the deep learning methods as well as in ablation experiments. The algorithm is suitable for solving user selection problem given a budget constraint. 

The problem maximizes return subject to a cost budget $B>0$:
\begin{align} 
\label{eq:pstatement_constrained} 
\begin{split}
  & \text{maximize} \quad \sum_{i=1}^n\tau^{*r}(\mathbf{x}^{(i)})z^{(i)} \\ 
  & \text{subject to} \quad \sum_{i=1}^n\tau^{*c}(\mathbf{x}^{(i)})z^{(i)} \leq B \\
	&z^{(i)} \in\{0, 1\} \quad \text{relaxed to} \quad 0\leq z^{(i)} \leq 1,
\end{split}
\end{align} 
where $z^{(i)}$ represents whether user $i$ is offered reward during a campaign and $B$ is the cost constraint. We denote treatment effect functions as $\tau^{*r}(\mathbf{x}^{(i)})$, $\tau^{*c}(\mathbf{x}^{(i)})$ as before. 

We provide a brief overview for solving Problem~\ref{eq:pstatement_constrained}. See Appendix \ref{apA} for a simplified derivation of this algorithm where we fit only one $\tau$ function.

We first use R-learner~\cite{nie2017quasi} to fit the $\tau^*$ functions to the training data. The solutions are deterministic. After applying the Lagrangian multiplier \(\lambda\), we have the Lagrangian for Problem~\ref{eq:pstatement_constrained}: 
\begin{equation}
\label{eq:duality_rlearner_langrangian}
  L(\mathbf{z}, \lambda)=-\sum_{i=1}^n\tau^{*r}(\mathbf{x}^{(i)})z^{(i)} + \lambda (\sum_{i=1}^n\tau^{*c}(\mathbf{x}^{(i)})z^{(i)} - B).
\end{equation}
The optimization in problem~\ref{eq:pstatement_constrained} can then be rewritten in its Dual form to maximize the Lagrangian dual function $g = \inf_{\mathbf{z}\in\emph{D}}L(\mathbf{z}, \lambda)$: 
\begin{equation}
\label{eq:duality_problem}
\max\limits_{\lambda} \inf_{\mathbf{z} \in \emph{D}} L(\mathbf{z}, \lambda) \quad
  \text{subject to} \ 0\leq z^{(i)} \leq 1, \lambda \geq 0.
\end{equation}
We iteratively solve for $z^{(i)}$ and $\lambda$ based on the two steps below.
\textbf{[1] Optimize $\mathbf{z^{(i)}}$:} Keeping $\lambda, \mathbf{\tau}$ fixed, we reduce problem~\ref{eq:duality_problem} to: 
\begin{equation} 
\begin{split}
  \label{eq:lagrangian_reduced}
  \text{maximize}\quad \sum_{i=1}^nz^{(i)} s^{(i)} \\
  \text{subject to} \quad 0\leq z^{(i)} \leq 1,
\end{split}
\end{equation} 
where we define the \emph{effectiveness score} $s^{(i)} = \tau^{*r}(\mathbf{x}^{(i)})-\lambda\tau^{*c}(\mathbf{x}^{(i)})$. It has a straightforward solution: assign $z^{(i)} = 1$ when $s^{(i)} \geq 0$; assign $z^{(i)} = 0$ when the ranking score $s^{(i)} < 0$. 

\textbf{[2] Optimize $\lambda$:} We then update $\lambda$. $\alpha$ is the learning rate. 
\begin{equation}
  \label{eq:update_lambda_algorithms_sec}
  \lambda\rightarrow \lambda + \alpha(B-\sum_{i=1}^n\tau^{*c}(\mathbf{x}^{(i)})z^{(i)}).
\end{equation}





We iteratively solve the Duality R-learner algorithm, where the optimal user selection \(z^{(i)}\) is re-computed at each point. The algorithm can jointly maximize return and minimize cost by directly solving the constrained optimization problem for balanced effectiveness. 

\subsection{Evaluation Methodology} 
\label{sec:evaluation}
The business objective is to achieve most incremental user revenue or engagement growth with a given cost budget. The revenue and cost here are two critical values to trade-off. 

\begin{figure}[h] 
  \centering 
  \includegraphics[width=0.75\linewidth]{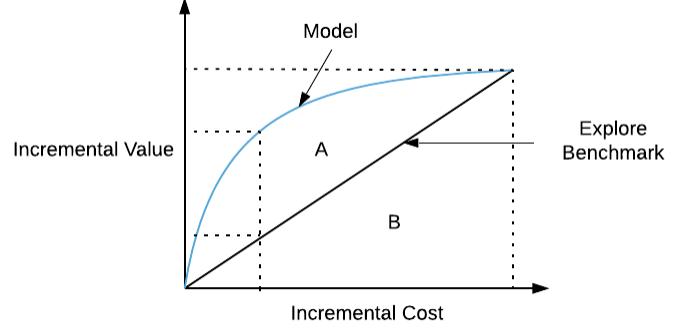} 
  \caption{Illustration of the Cost-Curve.} 
  \label{cost_curve_illustration} 
\end{figure}

\textbf{\emph{Cost Curve}}. With two treatment outcome $\tau^{r}$ and $\tau^{c}$, we draw a curve and use cost as X-axis and revenue as Y-axis shown in Fig.~\ref{cost_curve_illustration}. 
Samples are ordered by the effectiveness score $s^{(i)} = f(\mathbf{x}^{(i)})$ on the cost curve. For each point on the curve, we take the number of treatment samples at this point on the curve, multiplied by ATE (Average Treatment Effect) of this group. 
\begin{align*}
    \#\{T^{(i)} = 1 | s^{(i)} > s_i^{pth}\} \times ATE(x^{(i)} | s^{(i)} > s_i^{pth}).
\end{align*}

Each point represents aggregated incremental cost and value, usually both increasing from left to right. From origin to right-most of the curve, points on the curve represents the outcome if we include $q\%$ of the population for treatment, $q\in[0, 100]$.


If the score $s$ is randomly generated, the cost curve should be a straight line. If the score is generated by a good model, then the curve should be above the benchmark line, meaning for the same level of incremental cost, instances are selected to achieve higher incremental value. 

\textbf{\emph{Area Under Cost Curve (AUCC)}}. Similar to Area Under Curve of ROC curve, we define the normalized area under cost curve as  the area under curve divided by the area of rectangle extended by maximum incremental value and cost, or the area ratio $\frac{A + B}{2B}$. A and B are the area shown in the cost curve figure. The larger the AUCC, generally better the model. 

\section{Empirical Results} 
\label{sec:empirical_results} 

In this section, we compare proposed algorithms with prior art (Causal Forest, R-learner) on marketing and pubic datasets evaluated based on \ref{sec:evaluation}. We first describe the experiment setup and data, followed by analysis. In summary, our proposed deep learning methods perform significantly better than previous methods. 

\subsection{Experiments} 
The goal of our model is to rank users from most effective to least effective, so that the overall market-wide metrics are optimized.

\subsubsection{Experiment with Marketing Data} 
\label{sec:explore_exploit_exp} 
We adopt an explore and exploit experimental set-up, in the paradigm of reinforcement learning~\cite{liu2018explore} and multi-armed bandits~\cite{katehakis1987multi,li2010contextual,joachims2018deep}. We launch experiment algorithms in a cyclic fashion. For each cycle we have $2$ experiments: explore and exploit, which contain non-overlapping sets of users. \emph{For the production model investigated in this paper, we state this is possible and we follow the assumption that the explore treatment can be and is given randomly. The random treatment explore data is used for model training.} On the other hand, in exploit phase, we apply model to make treatment decisions and seek the best performance. The experiment design is illustrated in the following chart.  
\begin{figure}[h] 
  \centering 
  \includegraphics[width=\linewidth]{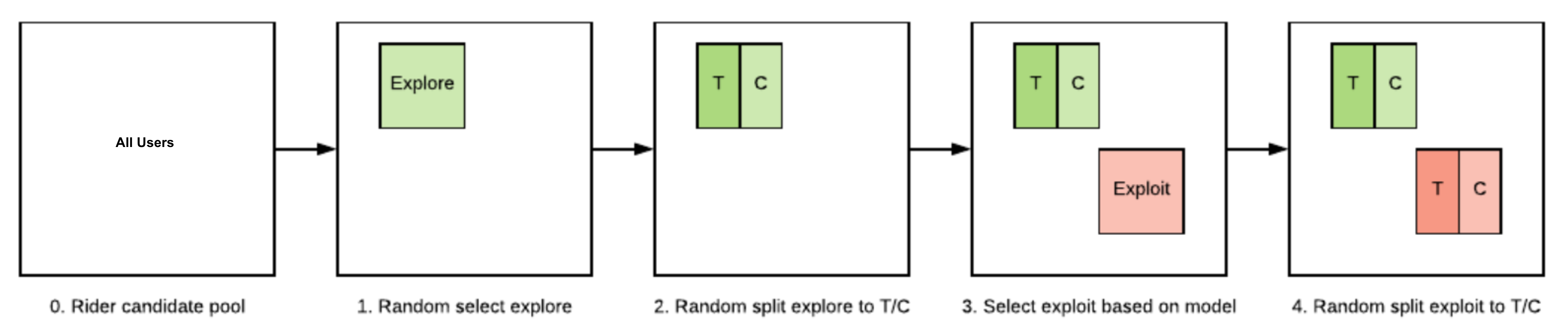} 
  \label{explore_exploit_chart} 
\end{figure} 

\emph{Explore}. Users are randomly selected into explore experiments from a larger candidate pool. This allows us to collect an unbiased dataset representing the entire population. Once the user pool is determined, we assign treatment / control with a fixed probability\footnote{The number of samples in explore is solely determined by the budget.}. 

\emph{Exploit}. We exclude users already in explore experiments in exploit experiment. Based on the available budget, we determine the cut-off percentage for user selection using scores from the trained model. Users above this threshold are assigned to the exploit group, which is used for product deployment and evaluated for online performance. 


 
We collect data from experiments following this design. For each sample, we will log their feature, experiment label (explore or exploit), treatment control assignment and outcomes (value and cost). Outcomes are aggregated within the experiment time-window. 

\textbf{Marketing Data.} To obtain data for model training and offline evaluation, we utilize a randomized \emph{explore} online experiment. We first randomly allocate users to control and treatment cohorts (A/B). For the treatment cohort, we give all users treatment.  In this experiment we collected millions of user level samples. Table~\ref{tab:example_data} shows illustrative data for the dataset we collected. 

\begin{table} 
  \caption{Example marketing dataset}
  \label{tab:example_data}
  \begin{tabular}{llllll}
    \toprule
    user id & strategy & $X_i$ & $T_i$& $Y^r$ & $Y^c$ \\
    \midrule
    A & explore & $(1.2, 3, ...)$ & $1$& $3$ & $2.3$ \\
    \midrule 
    B & exploit & $(2.4, 1, ...)$ & $0$& $1$ & $0.1$ \\ 
    \bottomrule
\end{tabular}
\end{table}




\subsubsection{Experiment with Public Datasets} We use two public datasets to experiment with our proposed methods. Detailed descriptions of data processing can be found in Appendix \ref{dataset}.

\textbf{US Census Data.} Our processed US Census Dataset has $46$ dimensions and $225814$ user samples. We select $T$ label as whether the person works more hours than the median of everyone else. $\tau_r$, the gain dimension, is ‘dIncome1’, and $\tau_c$, the cost dimension, is $-1.0$ multiplied with the number of children (‘iFertil’).

\textbf{Covertype Data.} Covertype Data is processed to have $51$ feature dimensions and $244365$ samples. We select $T$ treatment table to be whether the forest is closer to hydrology than the median of the filtered data. The gain outcome $\tau_r$ is whether the distance to wild fire is smaller than median and cost outcome $\tau_c$ is a penalty of selected tree types.

\subsubsection{Model Implementation} We briefly describe below. Detailed descriptions of model implementation can be found in Appendix \ref{model_config}.

\emph{Quasi-oracle estimation (R-learner)}. We use Linear Regression\footnote{Using SKLearn library's Ridge Regression with $0.0$ as the regularization weight.} as the base estimator. Since we need to define one CATE function to rank users, we use the R-learner to model the gain value incrementality $\tau_r$. 

\emph{Causal Forest}. To rank users with respect to cost vs. gain effectiveness, we estimate the conditional treatment effect function both for gain ($\tau_r$) and cost ($\tau_c$). We train two Causal Forest models. For evaluation, we compute the ranking score according to $\frac{\tau_r}{\tau_c}$.

\emph{R-learner with Multi-layer Perceptron}. To study the effect of using a deep learning model with R-learner, we replace linear regression model in R-learner with a two-layer neural network. 

\emph{Duality R-learner}. Similar to R-learner, we use Ridge Regression as the base estimator and constant propensity, and apply the model stated in Eq.~\ref{eq:lagrangian_score} for ease of online deployment. The iterative process to solve $\lambda$ in Eq.~\ref{eq:update_lambda_algorithms_sec} is less efficient. Since Ridge Regression is lightweight to train, in practice, we take the approach to select $\lambda$ with best performance on the validation set.

\emph{Direct Ranking}. To align with baseline and other methods in our experiments, we use a one layer network $\tanh(\mathbf{w}^T \mathbf{x} + b)$ as the scoring function, without weight regularization, the objective function (Eq.~\ref{eq:drm_deep_learning_summary})~\footnote{For numerical stability and differentiability, we apply a rectified activation function $\sigma(\cdot)$ (such as softplus function) to the denominator of Eq.~\ref{eq:drm_deep_learning_summary}).} stated in the algorithm section are used. We use the Adam optimizer with learning rate $0.001$ and default beta values. We compute gradients for entire batch of data, and run for $1500$ iterations.

\emph{Constrained Ranking}. We experiment with the Constrained Ranking Model with percentage constraints. We use a consistent percentage target at $40\%$. We apply a starting sigmoid temperature of $0.5$, and use annealing to increase temperature by $0.1$ every $10$ steps of Adam optimizer. The annealing schedule is obtained by cross-validation. 


\vspace{-0.2cm}
\subsection{Results on Causal Learning Models} 
\begin{figure}[h] 
  \includegraphics[width=0.85\linewidth]{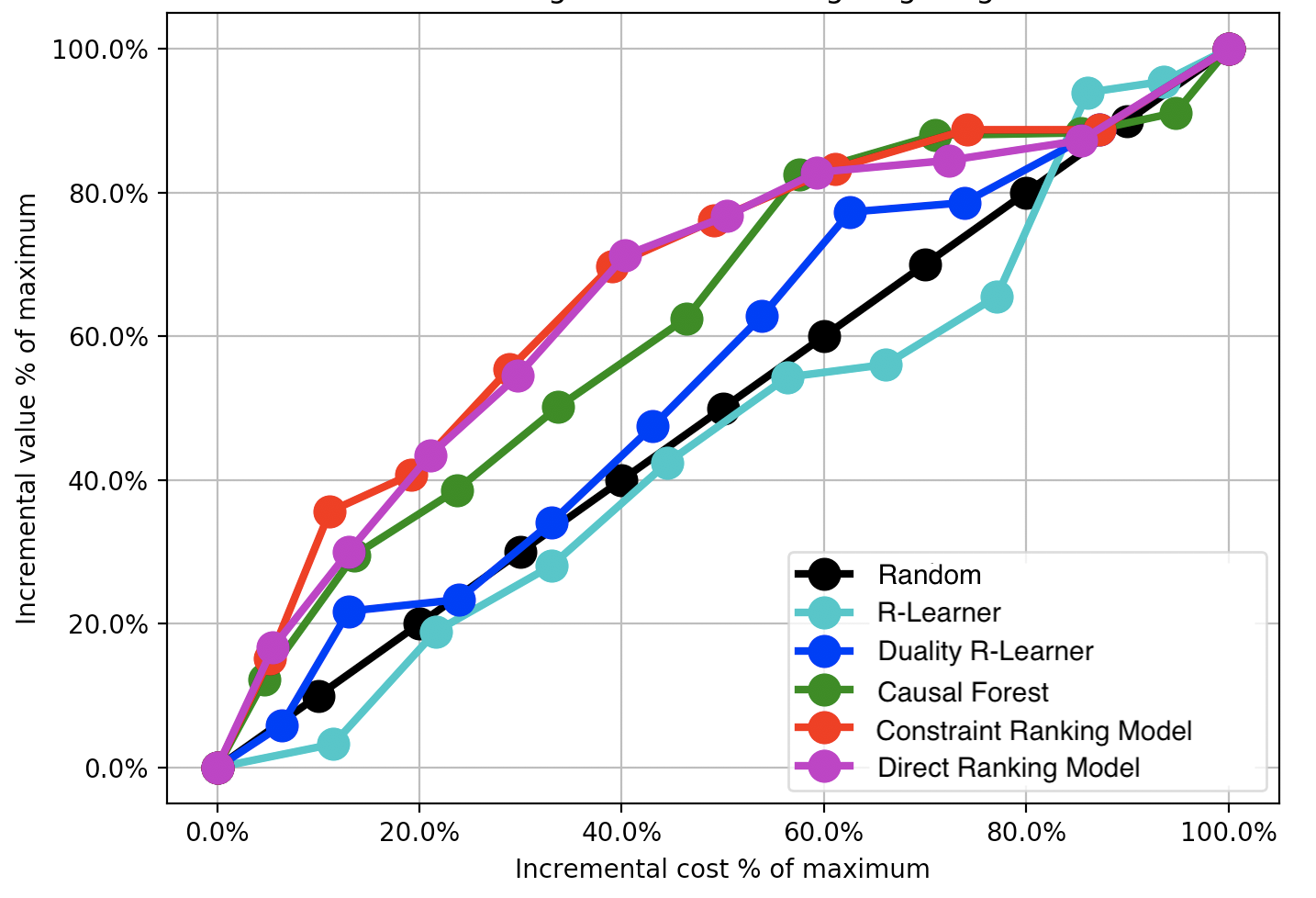} 
  \caption{Cost-Curve results for marketing data.} 
  \label{marketing_data_result} 
\end{figure}

Fig.~\ref{marketing_data_result} shows the cost curves for each model on marketing data test set. The baseline R-learner optimized for incremental gain $\tau_r$ could not account for the cost outcome and under-performs on our task. To be more fair, we use Duality R-learner as a benchmark for the marketing dataset. Causal Forest performs reasonably well. Direct Ranking out-performs previous models with \emph{22.1\%} AUCC improvement upon Duality R-learner, and Constrained Ranking algorithm is the best performing model on the marketing dataset, out-performing Duality R-learner by \emph{24.6\%} in terms of AUCC.

\begin{figure}[h] 
  \includegraphics[width=0.90\linewidth]{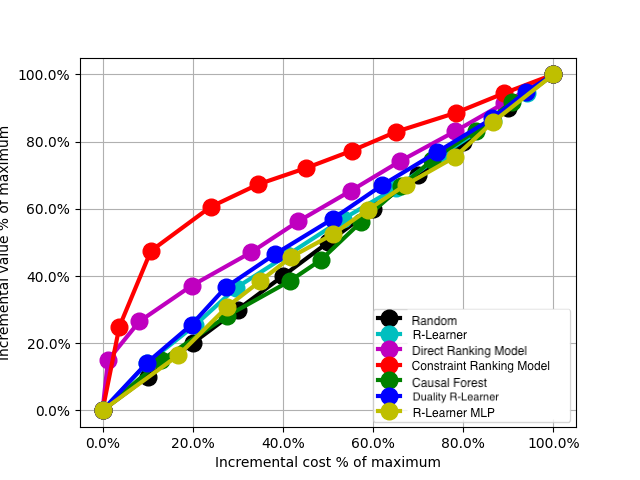} 
  \caption{Cost-Curve results for public US Census data.} 
  \label{fig:uscensus_result} 
\end{figure}

Fig.~\ref{fig:uscensus_result} shows results of causal models on US Census data. Duality R-learner works reasonably well. Direct Ranking and Constrained Ranking out-perform Duality R-learner by \emph{10.8\%} and \emph{31.2\%}, respectively with AUCC \emph{0.60} and \emph{0.72}. We show R-learner with MLP performs on par with Duality R-learner.

\begin{figure}[h] 
  \includegraphics[width=0.90\linewidth]{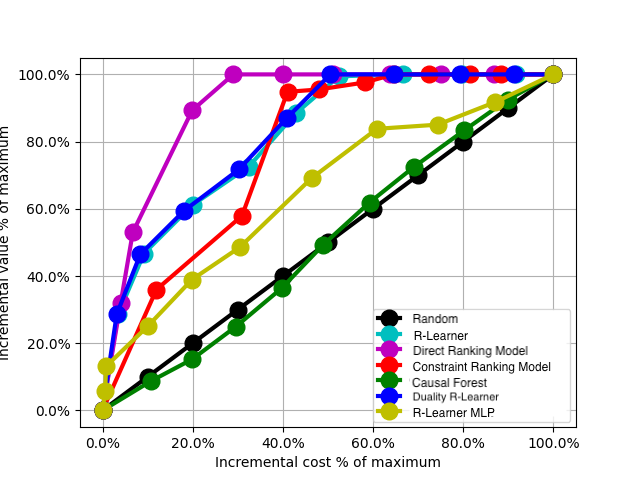} 
  \caption{Cost-Curve results for public Covtype data.} 
  \label{fig:covtype_result}
\end{figure}

Fig.~\ref{fig:covtype_result} shows results on Covtype data. Direct Ranking out-performs Duality R-learner by \emph{9.9\%} with AUCC \emph{0.907}. The Constrained Ranking algorithm performs less well on this data, likely due to the already high AUCC and good performance of baseline models even at low percentages. The R-learner with MLP does not yield better result.


In Summary, Table~\ref{tab:summary_result_table} shows that Constrained Ranking and Direct Ranking algorithms perform consistently better than R-learner and Duality R-learner, often by more than \emph{10\%}. At $20\%$ of total incremental cost on marketing and US census data, the DRM model offers around $2$x more incremental gain than most existing models. 

We attribute this performance to the model's ability to aggregate multiple treatment effect functions into a combined objective, and optimizing all the parameters jointly as illustrated in Fig.~\ref{fig:algorithm_illustration}. As an ablation, the R-learner with a deep model MLP do not reach similar good performance, and often perform worse than linear R-learner. The experiments verify our proposal that the deep learning approach such as DRM and Constraint Ranking models, as compared to the greedy approach~\cite{nie2017quasi,kunzel2017meta}, can significantly improve empirical performance on user growth marketing and public datasets. 
\begin{table}
  \caption{Summary of AUCC results across models and datasets.} 
  \label{tab:summary_result_table}
  \begin{tabular}{llllll}
    \toprule
    Algorithm & Mark. &\% imp. & USCensus &Covtype\\ 
    \midrule 
    Random & 0.500 && 0.500 & 0.500 \\
    \midrule 
    R-learner G & 0.464 && 0.529 & 0.817 \\
    R-learner MLP &  & & 0.506 & 0.651 \\
    Duality R-learner & 0.544 &0.0\%& 0.545 & 0.825 \\
    Causal Forest & 0.628 &15.4\%& 0.499 & 0.498 \\
    Direct Ranking & 0.664 &22.1\%& 0.604 & \textbf{0.907} \\
    Constrained Ranking & \textbf{0.678} &24.6\%& \textbf{0.715} & 0.770 \\
    \bottomrule
\end{tabular}
\end{table} 

\textbf{Direct Ranking Model with Propensity}
We validate that model generalization on public datasets can be improved with the propensity weighting algorithm. 

We first learn a propensity function $e(\mathbf{x})$ with regression. Given user feature $\mathbf{x}$, this function estimates $P(T=1|\mathbf{x})$. Then, leveraging Eq.~\ref{eq:treament_effect_function_with_propensity}, we substitute the estimated values of $e(\mathbf{x})$ in the training set to form an objective function suitable for DRM. For propensity related experiments we use a more simple objective\footnote{The objective is: $\quad \text{maximize} \quad \tau^{*r} - \alpha \tau^{*c}$. We set $\alpha=1.3$ for R-learner with propensity and $\alpha=1.5$ for DRM with propensity.}.

Model generalization improves if treatment bias or propensity assumption exist both in the training and test sets. For public datasets, this holds true. To evaluate the model generalization, we use propensity weighted objective in Eq.~\ref{eq:drm_deep_learning_summary} as the evaluation metric, and compute it on the test set. ~\footnote{For marketing production use-case, we have full control of treatment application at test time. While for public datasets, we assume that we can't fully control treatment decisions at test time, and the treatment bias still exists in the test set. The DRM with propensity model produces rankings and selections meaningful in the counterfactual sense: e.g. if the top-ranking user had been given a different treatment, the overall objective would improve.}

For R-learner, we evaluate the same objective as Eq.~\ref{eq:drm_deep_learning_summary} on the test set. To compute the user effectiveness probabilities $p_i$, we take the uplift predictions from R-learner and feed through a softmax function for each cohort. 

Our experiments demonstrate that incorporating propensity function significantly enhances the test metric of both the R-learner and DRM when selecting users or samples. As shown in Table~\ref{tab:generalization_propensity_table}, the inclusion of propensity consistently improves the results across different selection percentages ($q$).

We emphasize the importance of theoretical generalization of the DRM model with propensity functions, and offer the theoretical derivation in Appendix~\ref{apB}. These results also show practical benefits of incorporating propensity, particularly in scenarios where counterfactual reasoning is desired and treatment decisions inherently exhibit propensity bias.

\begin{table}[h]
\caption{Applying propensity with DRM on public datasets}
    \centering
    \resizebox{\columnwidth}{!}{
        \begin{tabular}{|c|c|c|c|c|}
            \hline
            \multicolumn{5}{|c|}{\textbf{US Census: test set generalization scores}} \\
            \hline
            $q$ & R-learner & R-learner w. Propensity & DRM & DRM w. Propensity \\
            \hline
            15\% & 0.0265 & \textbf{0.0416} & 0.4026 & 0.3427 \\
            \hline
            20\% & 0.0234 & \textbf{0.0370} & 0.4426 & 0.3896 \\
            \hline
            30\% & 0.0185 & \textbf{0.0357} & 0.3216 & \textbf{0.4591} \\
            \hline
            40\% & 0.0149 & \textbf{0.0271} & 0.3148& \textbf{0.5151} \\
            \hline
            60\% & 0.0080 & \textbf{0.0129} & 0.3679 & \textbf{0.5684} \\
            \hline
            80\% & 0.0053 & \textbf{0.0099} & 0.4224 & \textbf{0.6117} \\
            \hline
            100\% & 0.0039 & \textbf{0.0072} & 0.5144 & \textbf{0.8085} \\
            \hline
            \multicolumn{5}{|c|}{\textbf{Covtype: test set generalization scores}} \\
            \hline
            q & R-learner & R-learner w. Propensity & DRM & DRM w. Propensity \\
            \hline
            15\% & 0.0247 & \textbf{0.0286} & 0.1364 & \textbf{0.5809} \\
            \hline
            20\% & 0.0180 & \textbf{0.0203} & 0.1447 & \textbf{0.6034} \\
            \hline
            30\% & 0.0114 & \textbf{0.0131} & 0.1563 & \textbf{0.6431} \\
            \hline
            40\% & 0.0086 & \textbf{0.0105} & 0.1543 & \textbf{0.6747} \\
            \hline
            60\% & 0.0056 & \textbf{0.0076} & 0.1610 & \textbf{0.7442} \\
            \hline
            80\% & 0.0043 & \textbf{0.0062} & 0.1148 & \textbf{0.8749} \\
            \hline
            100\% & 0.0035 & \textbf{0.0049} & 0.0810 & \textbf{1.1420} \\
            \hline            
        \end{tabular}
    }
    \label{tab:generalization_propensity_table}
\end{table}

\begin{figure}[h] 
  \centering 
  \includegraphics[width=\linewidth]{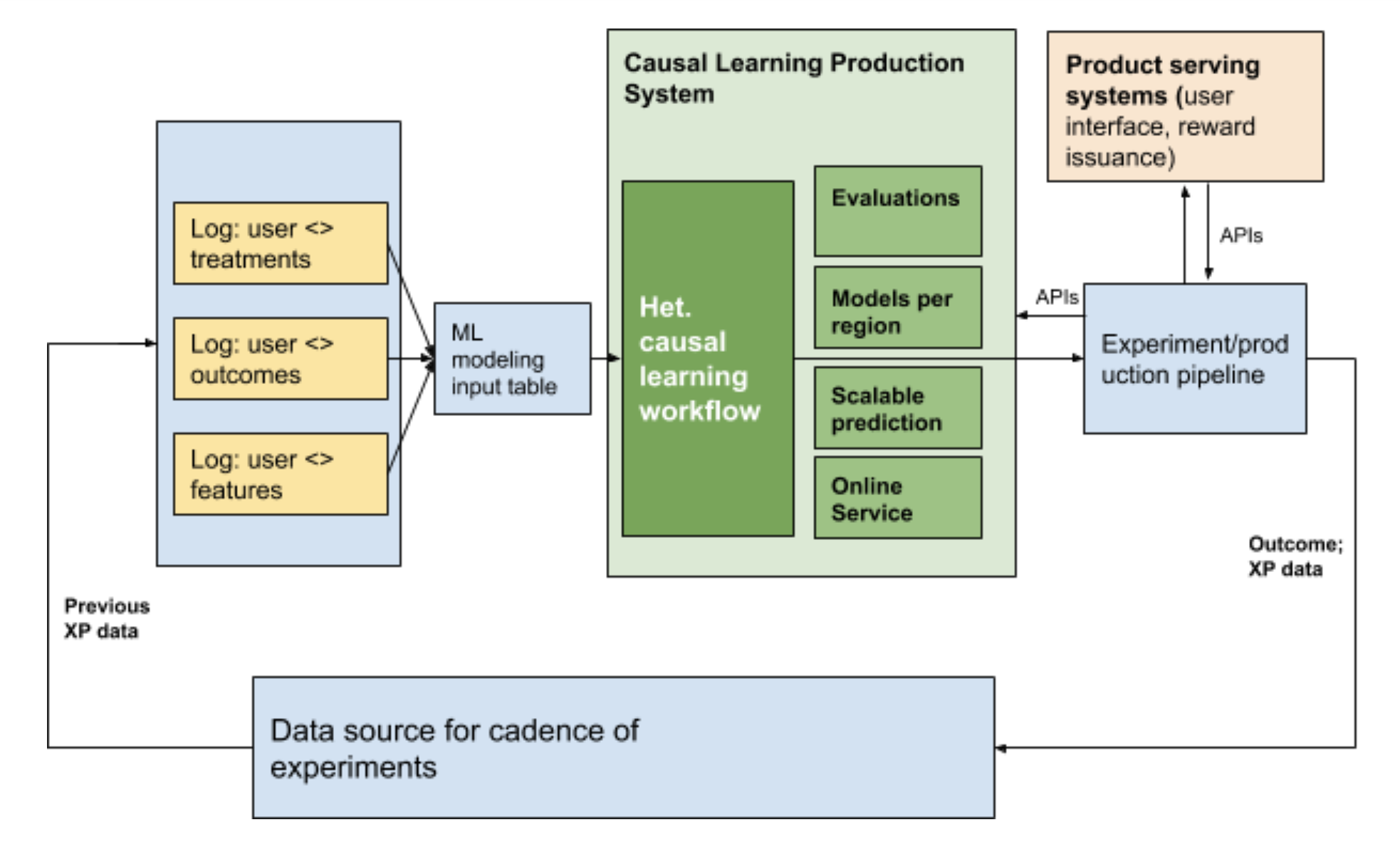} 
  \caption{Production system for causal learning.} 
  \label{fig:kdd_prod_eng} 
\end{figure}

\textbf{Description of the Production System.} The models we propose have been deployed in production and are operating across multiple regions. Unlike traditional machine learning systems, we have developed a Heterogeneous Causal Learning Workflow (HCLW) system to support the collection of user growth experiment data and observed outcomes. Data from previous production experiments provide training signals for subsequent launches, enabling the product model to improve its decisions iteratively across a sequence of launches. The design of this system is shown in Fig.~\ref{fig:kdd_prod_eng}. The data are collected from previous launches in the form of Table~\ref{tab:example_data}, and are stored in offline storage before feeding into the causal learning pipeline. The pipeline produces the trained model, evaluation, and service components. The service components use the trained model to generate user-level decisions, interact with launch pipelines, and connect with the product serving system through APIs to issue rewards directly to users via the product interface in production.

\emph{\textbf{Production and Offline Evaluation.}} We consider the alignment of evaluation results. Unlike the full cost-curve for offline evaluation, in online case, we could measure one specific point on the cost curve. The slope of the straight line between that point and origin measures the general cost effectiveness. This slope is given as $R$ in Eq.\ref{eq:slope_r} and it sufficiently captures the performance if both models have similar spend level. 

\begin{equation}
  \label{eq:slope_r}
  R = \frac{ATE^r(x^{(i)}|selected)}{ATE^c(x^{(i)}|selected)}.
\end{equation}

Our online setup includes both explore and exploit phases. Specifically, when comparing two models, such as DRM and Causal Forest, we configure one explore group (random selection) and two exploit groups (model-based selection). Within each selected group, users are then randomly split into treatment and control.

To ensure numerical metrics are comparable across these settings, we use $R$ for explore as benchmark We then compute the relative efficiency gain using Eq.~\ref{eq:ratio_r}, which quantifies the improvement achieved by the models relative to the benchmark.
\begin{equation}
  \label{eq:ratio_r}
  \frac{R_{exploit}-R_{explore}}{R_{explore}}.
\end{equation}


Online results are consistent with the offline results, which shows that the models perform significantly better than explore and DRM consistently out-performs quasi-oracle estimation (R-learner) and Causal Forest.

\section{Conclusion and Future Work} 
\label{sec:conclusion}

\subsection{Conclusion} 
We propose a novel user growth ranking method to optimize aggregated heterogeneous treatment functions. The method provides theoretical foundations on the functional forms of treatment effect complete with propensity weighting. It applies a deep learning approach to effectively learn the model parameters. In addition, we provide empirical evaluation metrics and compare various methods on marketing and public datasets. After successful test, this method has been deployed to production in many regions all over the world. Based on foundational research, our proposed algorithms and methods achieve significantly better performance than existing methods. 

\subsection{Future Work} 
\emph{Smart Explore/Exploit}. In current work we use epsilon-greedy explore, where we split a fixed percentage of budget to spend on fully randomized explore to collect data for model training. As a better approach, we will try to use multi-arm bandit~\cite{joachims2018deep} or Bayesian optimization framework to guide our smart explore based on the model uncertainty. 

\emph{Deep Embedding}. Raw time and geographical features are extremely sparse. Various embedding techniques have been used for sparse features while they are not used in treatment effect models. Now that there is a general loss layer to be incorporated with any deep learning structure, we can start to work on embeddings specifically for treatment effect models. 


\bibliographystyle{ACM-Reference-Format}
\bibliography{references}
\appendix

\section{Duality R-learner Algorithm} \label{apA}
We describe the duality method with Lagrangian multipliers to solve the constrained optimization problem for Eq. \ref{eq:constrained_problem}. 

Our objective is to maximize gain subject to a budget ($B>0$) constraint. To solve the problem, we relax $z^{(i)}\in\{0, 1\}$ variables to continuous ones, so the problem is defined as: 
\begin{align} 
\label{eq:constrained_problem} 
\begin{split}
  &\text{minimize} \quad -\sum_{i=1}^n\tau^{*r}(\mathbf{x}^{(i)}) \cdot z^{(i)} \\ 
  &\text{subject to} \quad \sum_{i=1}^n\tau^{*c}(\mathbf{x}^{(i)}) \cdot z^{(i)} \leq B \\
  & \text{where} \quad z^{(i)} \in\{0, 1\} \quad \text{relaxed to} \quad 0\leq z^{(i)} \leq 1.
\end{split}
\end{align} 
The variables $z^{(i)}$ represent whether we offer a reward to the user $i$ during a campaign and $B$ is the cost constraint.

We first fit the $\tau^*$ functions to the training data, by leveraging quasi-oracle estimation ~\cite{nie2017quasi}. 

Then, we have $\tau^{*r}(\mathbf{x}^{(i)})$ and $\tau^{*c}(\mathbf{x}^{(i)})$ given and fixed for each data point $\mathbf{x}^{(i)}$. We apply the Lagrangian multiplier \(\lambda\). The Lagrangian for problem \ref{eq:constrained_problem} is: 
\[
  L(\mathbf{z}, \lambda)=-\sum_{i=1}^n\tau^{*r}(\mathbf{x}^{(i)}) \cdot z^{(i)} + \lambda (\sum_{i=1}^n\tau^{*c}(\mathbf{x}^{(i)}) \cdot z^{(i)} - B). \\ 
\]
The optimization in problem~\ref{eq:constrained_problem} can then be rewritten in its Dual form to maximize the Lagrangian dual function $g = \inf_{\mathbf{z}\in\emph{D}}L(\mathbf{z}, \lambda)$: 
\begin{align*}
\max\limits_{\lambda} \inf_{\mathbf{z} \in \emph{D}} L(\mathbf{z}, \lambda) \quad
  \text{subject to} \ 0\leq z^{(i)} \leq 1, \lambda \geq 0.
\end{align*}

To solve this problem using duality, we must consider certain caveats and determine whether the dual and primal problems achieve the same minimum.
\begin{itemize} 
\item Given $p(\mathbf{z}, \lambda) = -\sum_{i=1}^n\tau^{*r}(\mathbf{x}^{(i)}) \cdot z^{(i)}$, we know, for the optimal values of the primal and dual problems, the relationship $p^* \leq g^*$ holds according to the principles of convex optimization. Equality $p^* = g^*$ holds if $p$ and $g$ are convex functions, and the \emph{Slater constraint qualification} is met, which requires the primal problem to be strictly feasible. 
\item To satisfy the Slater condition, there must exist a feasible point such that the inequality constraints are strictly satisfied. For any positive value $B > 0$, we can always satisfy the strict inequality  $\sum_{i=1}^n\tau^{*c}(\mathbf{x}^{(i)}) \cdot z^{(i)} < B$ by selecting sufficiently small values for certain \(z^{(i)}\). Furthermore, in contexts like marketing campaign, the value of $B$ is usually large, which supports that the Slater qualifications hold in this setting. 
\end{itemize} 

\noindent From the analysis above, problem~\ref{eq:constrained_problem} and its dual problem~\ref{eq:duality_problem} are equivalent, and we can solve problem~\ref{eq:duality_problem} by iteratively optimizing with respect to $\mathbf{z}$ and $\lambda$.  

\textbf{Optimize $\mathbf{z^{(i)}}$:} Keeping $\lambda, \mathbf{\tau}$ fixed, as $\lambda$ and $B$ are constants, we can turn problem~\ref{eq:duality_problem} into: 
\begin{align*} 
\begin{split}
  \text{maximize}\quad \sum_{i=1}^nz^{(i)} s^{(i)} \\
  \text{subject to} \quad 0\leq z^{(i)} \leq 1,
\end{split}
\end{align*} 
where we define the \emph{effectiveness score} $s^{(i)} = \tau^{*r}(\mathbf{x}^{(i)})-\lambda\tau^{*c}(\mathbf{x}^{(i)})$. This optimization problem has a straightforward solution: assign the multiplier $z^{(i)} = 1$ when the ranking score $s^{(i)} \geq 0$; assign $z^{(i)} = 0$ when the ranking score $s^{(i)} < 0$. 

\textbf{Optimize $\lambda$:} We take the derivative of $L$ with regard to $\lambda$, $\frac{\partial g}{\partial \lambda}=B-\sum_{i=1}^n\tau^{*c}(\mathbf{x}^{(i)}) \cdot z^{(i)}$. We then update $\lambda$ by Eq. \ref{eq:update_lambda} where $\alpha$ is the learning rate. 
\begin{equation}
  \label{eq:update_lambda}
  \lambda\rightarrow \lambda + \alpha(B-\sum_{i=1}^n\tau^{*c}(\mathbf{x}^{(i)}) \cdot z^{(i)}).
\end{equation}
Based on the two steps above, we can iteratively solve for both $z^{(i)}$ and $\lambda$ \cite{bertsekas1999nonlinear} .

The above constrained optimization algorithm is implemented for our experiments. For a more simplified \emph{Duality R-learner} algorithm, we also implemented an approach to combine the two $\tau^*$ functions into one model. Instead of learning $\tau^{*r}$ and $\tau^{*c}$ respectively, we fit a single \emph{scoring model} $s^{(i)}=\tau^{*E}(\mathbf{x}^{(i)})$ in Eq. \ref{eq:lagrangian_score}. Note that the Duality solution suggests we should include any sample with $\hat\tau^{*E}(x^{(i)})>0$. The larger $\hat\tau^{*E}(x^{(i)})$ is, the more contribution the sample \(i\) will have and thus the higher ranking it should get. 

We define \(s^{(i)}\) to be:
\begin{equation} 
  \label{eq:lagrangian_score} 
  s^{(i)}=\tau^{*E}(\mathbf{x}^{(i)})=\tau^{*r}(\mathbf{x}^{(i)})-\lambda\tau^{*c}(\mathbf{x}^{(i)}) .
\end{equation} 
This form is linear, so we can use $Y^E=Y^r-\lambda Y^c$ instead of the the original $Y$ (single outcome for value and cost respectively) in the estimators above. Specifically, 
\begin{align*} 
  \tau^{*E} &= \tau^{*r}(\mathbf{x}) - \lambda \tau^{*c}(\mathbf{x}) \\
  &= E((Y^r_1 - \lambda Y^c_1 - (Y^r_0  - \lambda Y^c_0)| \mathbf{X} = \mathbf{x}) \\
  &= E(Y^E_1 - Y^E_0 | \mathbf{X} = \mathbf{x}) .
\end{align*} 
We then train the regression model through the quasi-oracle estimation method, and the output $\tau^{*E}$ could be used directly. This has two benefits: first, we optimize a joint model across $Y^r$ and $Y^c$ for the parameters to be able to find correlations jointly; second, for production and online service, we arrive at one single model to perform prediction. 

With the trained regression model, we are able to compute $\tau^{*r}$ and $\tau^{*c}$ for each user \(i\). We then re-compute \(z^{(i)}\) using the dual problem. This iterative process enables convergence toward an optimal solution for user selection and resource allocation.

\section{Propensity Score Derivation for DRM}
\label{apB}
As discussed in Eq. \ref{eq:pstatement_unconstrained_opt}, our aim is to minimize the cost per unit gain. For each user with feature $x$, $\tau^{*r}(\mathbf{x}) = E(Y_1^r - Y_0^r | \mathbf{X} = \mathbf{x})$ and cost $\tau^{*c}(\mathbf{x}) = E(Y_1^c - Y_0^c | \mathbf{X} = \mathbf{x})$ are expectations across selected user portfolios, where $Y_1^r$ and $Y_1^c$ are the reward and cost, respectively, for a treated user and $Y_0^r$ and $Y_0^c$ are the reward and cost for an untreated user.

Without loss of generality, we derive uplift $\tau$ without the subscript as it could be extended to $\tau^r$ and $\tau^c$ as below.
\ignorespacesafterend

We start with the fundamental definition of treatment effect:
\begin{align*}   
\tau&{*} = E(Y_1 - Y_0) = E(Y_1) - E(Y_0),
\end{align*}
where 
\begin{align*}
E(Y_1) &= E\left(\frac{Y_1}{e(\mathbf{x})} e(\mathbf{x})\right) \nonumber \\
       &= E\left(\frac{Y_1}{e(\mathbf{x})}E(T | \mathbf{X}=\mathbf{x})\right) \nonumber\\
       &= E\left(\frac{Y_1}{e(\mathbf{x})}E(T | \mathbf{X}=\mathbf{x}, Y_1)\right)\nonumber \\
       &= E\left(E\left(\frac{Y_1T}{e(\mathbf{x})} \middle| X, Y_1\right)\right) \nonumber\\
       &= E\left(\frac{Y_1T}{e(\mathbf{x})}\right),
\end{align*}
where the third step follows from unconfoundedness assumption~\cite{lunceford04stratification,nie2017quasi}. 

Similarly, $E(Y_0) = E(\frac{Y_0 (1-T)}{1-e(\mathbf{x})})$.  Therefore, combining the previous steps, we arrive at: \begin{align*} 
\begin{split} 
\tau^{*} &= E(Y_1 - Y_0) = E(Y_1) - E(Y_0) \\
&= E(\frac{Y_1T}{e(\mathbf{x})})  - E(\frac{Y_0 (1-T)}{1-e(\mathbf{x})}) \\ 
&= E(E(\frac{Y_1T}{e(\mathbf{x})}|T)) - E(E(\frac{Y_0(1-T)}{1-e(\mathbf{x})}|T)), \\ 
\end{split}
\end{align*} 
where the last step follows from the law of total expectation.

We can then expand the equation by definition of expectation: 
\begin{align*} 
    \begin{split} 
    \bar{\tau}^{*} &= P(T = 1) E(\frac{Y_1T}{e(\mathbf{x})}|T = 1) + \cancel{P(T = 0) E(\frac{Y_1T}{e(\mathbf{x})}| T = 0)}\\
    &\quad - \cancel{P(T = 1) E(\frac{Y_0 (1-T)}{1-e(\mathbf{x})}|T = 1)} - P(T = 0) E(\frac{Y_0 (1-T)}{1-e(\mathbf{x})} | T = 0) .
    \end{split} 
\end{align*} 
For each user, we use $p_i$ as a relaxation of the binary decision variable $z^{(i)}$ in the Duality R-Learner problem in Appendix \ref{dual_r_learner}, the effectiveness weighting to select user $i$. The higher the $p_i$, the higher likelihood of selecting the sample, and we constrain $\sum p_{i, T_i = 0} = 1$, $\sum p_{i, T_i=1} = 1$, i.e. the effectiveness measures for users in the treatment cohort, and control cohort sums to $1$ respectively. This normalization gives the effective measure a probabilistic interpretation. We define $\hat{e}$ to be the overall propensity, or probability of treatment, estimated by mean of the binary treatment labels across users. For each user, the $e(\mathbf{x})$ term is estimated from a pre-trained propensity function using logistic regression. Then, we expand the expectation terms $E(\frac{Y_1T}{e(\mathbf{x})}|T = 1)$, $E(\frac{Y_0 (1-T)}{1-e(\mathbf{x})} | T = 0)$ using the normalized effectiveness weighting, potential outcomes $Y^{(i)}_1$, $Y^{(i)}_0$ and observed outcome labels $Y^{(i)}$: 

\begin{align} 
\label{eq:drm_propensity_3} 
\begin{split} 
\tau^{*} &= \hat{e} \sum_{T_i =1} \frac{Y^{(i)}_1 p_{i}}{e(\mathbf{x}^{(i)})} - (1-\hat{e}) \sum_{T_i = 0} \frac{Y^{(i)}_0 p_i}{1-e(\mathbf{x}^{(i)})} \\ 
& = \hat{e} \sum_{i=1}^n \frac{1}{e(\mathbf{x}^{(i)})} Y^{(i)}_1 p_{i} \mathbb{I}_{T_i=1} - (1-\hat{e}) \sum_{i = 0}^n \frac{1}{1-e(\mathbf{x}^{(i)})} Y^{(i)}_0 p_i \mathbb{I}_{T_i=0} \\ 
& = \hat{e} \sum_{i=1}^n \frac{1}{e(\mathbf{x}^{(i)})} Y^{(i)} p_{i} \mathbb{I}_{T_i=1} - (1-\hat{e}) \sum_{i = 0}^n \frac{1}{1-e(\mathbf{x}^{(i)})} Y^{(i)} p_i \mathbb{I}_{T_i=0}.
\end{split} 
\end{align} 
Eq.~\ref{eq:drm_propensity_3} is a generalized form of the $\tau^{*}$ including part of the objective function with propensity weighting. When $\hat{e} = e(\mathbf{x}^{(i)})$ is constant in a fully randomized experiment, the term becomes: 
\begin{align*} 
  \bar\tau^{*}=\sum_{i=1}^n Y^{(i)}p_i(\mathbb{I}_{T_i=1} - \mathbb{I}_{T_i=0}).
\end{align*} 

For numerical stability and differentiability, we apply a rectified activation function $\sigma_r(\cdot)$ (such as softplus function), leading to the final form in Eq.~\ref{eq:propensity_drm_objective}.

\begin{equation}
\small
\label{eq:propensity_drm_objective}
\frac{\tau^{*c}}{\tau^{*r}} =
\frac{\sigma_r \left(\hat{e} \sum\limits_{i=1}^{n} \frac{1}{e(\mathbf{x}^{(i)})} Y^{c(i)} p_i \mathbb{I}_{T^{(i)}=1} - (1-\hat{e}) \sum\limits_{i=1}^{n} \frac{1}{1-e(\mathbf{x}^{(i)})} Y^{c(i)} p_i \mathbb{I}_{T^{(i)}=0} \right)}
{\sigma_r \left(\hat{e} \sum\limits_{i=1}^{n} \frac{1}{e(\mathbf{x}^{(i)})} Y^{r(i)} p_i \mathbb{I}_{T^{(i)}=1} - (1-\hat{e}) \sum\limits_{i=1}^{n} \frac{1}{1-e(\mathbf{x}^{(i)})} Y^{r(i)} p_i\mathbb{I}_{T^{(i)}=0} \right)}.
\end{equation}

This formulation enables counterfactual learning by integrating inverse propensity scoring into the optimization framework, ensuring an unbiased and stable estimation of aggregated effectiveness.

\section{Causal Experiments Designed with Public Datasets} 
\label{dataset}

We experiment on public datasets to validate our proposed methodologies.

\textbf{US Census 1990} The US Census (1990) Dataset (Asuncion \& Newman, 2007~\cite{uscensulink} contains data for people in the census. Each sample contains a number of personal features (native language, education...). The features are pre-screened for confounding variables, we left out dimensions such as other types of income, marital status, age and ancestry. This reduces features to d = 46 dimensions. Before constructing experiment data, we first filter with several constraints. We select people with one or more children (\emph{`iFertil'} $\geq$ 1.5), born in the U.S. (\emph{`iCitizen'} = 0) and less than 50 years old (\emph{`dAge'} $<$ 5), resulting in a dataset with $225814$ samples. We select `treatment' label as whether the person works more hours than the median of everyone else, and select the income (\emph{`dIncome1'}) as the gain dimension of outcome for $\tau_r$, then the number of children (`iFertil') multiplied by $-1.0$ as the cost dimension for estimating $\tau_c$. The hypothetical meaning of this experiment is to measure the cost effectiveness, and evaluate who in the dataset is effective to work more hours.

\textbf{Covertype Data} The Covertype Dataset (Asuncion \& Newman, 2007) contains the cover type of northern Colorado forest areas with tree classes,  distance to hydrology, distance to wild fire ignition points, elevation, slope, aspect, and soil type. We pre-filter and only consider two types of forests: \emph{`Spruce-Fir'} and \emph{`Lodgepole Pine '}, and use data for all forests above the median elevation. This results in a total of $244365$ samples. After processing and screening for confounding variables, we use 51 features for model input. With the filtered data, we build experiment data by assuming we are able to re-direct and create water source in certain forests to fight wild fires. Additionally, we would like to ensure the covertype trees are balanced by changing the hydrology with preference to `Spruce-Fir'. Thus, the treatment label is selected as whether the forest is close to hydrology, concretely, distance to hydrology is below median of the filtered data. The gain outcome is a binary variable for whether distance to wild fire points is smaller than median, and cost outcome is the indicator for `Lodgepole Pine' (1.0, undesired) as opposed to `Spruce-Fir' (0.0, desired). 

Marketing data, public US Census and Covtype datasets are split into 3 parts: train, validation and test sets with respective percentages 60\%, 20\%, 20\%. We use train and validation sets to perform hyper-parameter selection for each model type. The model is then evaluated on the test set.

\section{Model Configuration}
\label{model_config}
We perform hyper-parameter search and arrive at the best sets of hyper-parameters for several models.

\emph{Causal Forest} For hyper-parameters, we perform search on deciles for parameters \emph{num\_trees, min.node.size}, and at \emph{0.05} intervals for \emph{alpha, sample.fraction parameters}. We also leverage the \emph{tune.parameters} option for the grf package. Best parameters are the same for all three datasets we experimented: \emph{num\_trees}$=100$ (50 trees for each of the two CATE function, $\tau_r$, $\tau_c$), \emph{alpha}$=0.2$, \emph{min.node.size}$=3$, \emph{sample.fraction}$=0.5$

\emph{R-learner with MLP} With validation results, we find the optimal number of hidden layers to be 92 for US Census and 100 for Covertype data.

\emph{Duality R-learner} We determine the value of $\lambda$ in the model  through hyper-parameter search on deciles and mid-deciles, e.g. $\lambda\in \{0.001, 0.005, 0.01, 0.05\}$; best $\lambda$ for marketing data is $0.1$, for US Census and Covertype data is $0.05$.

\end{document}